\documentclass[letterpaper]{article} 
\usepackage{aaai2026}  
\usepackage{times}  
\usepackage{helvet}  
\usepackage{courier}  
\usepackage[hyphens]{url}  
\usepackage{graphicx} 
\usepackage{natbib}  
\usepackage{caption} 
\usepackage{algorithm}
\usepackage{algorithmic}
\usepackage{tabularx}
\usepackage{booktabs}
\usepackage{multirow}   
\usepackage{makecell} 
\usepackage[table]{xcolor}
\usepackage{amsmath}
\usepackage{newfloat}
\usepackage{listings}
\DeclareCaptionStyle{ruled}{labelfont=normalfont,labelsep=colon,strut=off} 
\floatstyle{ruled}
\newfloat{listing}{tb}{lst}{}
\floatname{listing}{Listing}
\title{SymGS : Leveraging Local Symmetries for 3D Gaussian Splatting Compression}

\author {
    Keshav Gupta\equalcontrib \textsuperscript{\rm 1,\rm 3},
    Akshat Sanghvi\equalcontrib \textsuperscript{\rm 1},
    Shreyas Reddy Palley\textsuperscript{\rm 1},
    Astitva Srivastava\textsuperscript{\rm 1},
    Charu Sharma\textsuperscript{\rm 1},
    Avinash Sharma\textsuperscript{\rm 2}
}
\affiliations{
    \textsuperscript{\rm 1}IIIT Hyderabad\\
    \textsuperscript{\rm 2}IIT Jodhpur\\
    \textsuperscript{\rm 3}University of California, San Diego\\
    \{keshav.gupta,
    akshat.sanghvi,
    shreyas.palley\}@students.iiit.ac.in,
    astitva.s@research.iiit.ac.in,
    charu.sharma@iiit.ac.in,
    avinashsharma@iitj.ac.in
}

\usepackage{bibentry}

\begin{document}



\twocolumn[{%
\renewcommand\twocolumn[1][]{#1}%

\maketitle

\begin{center}
    \centering
    \includegraphics[width=\textwidth,]{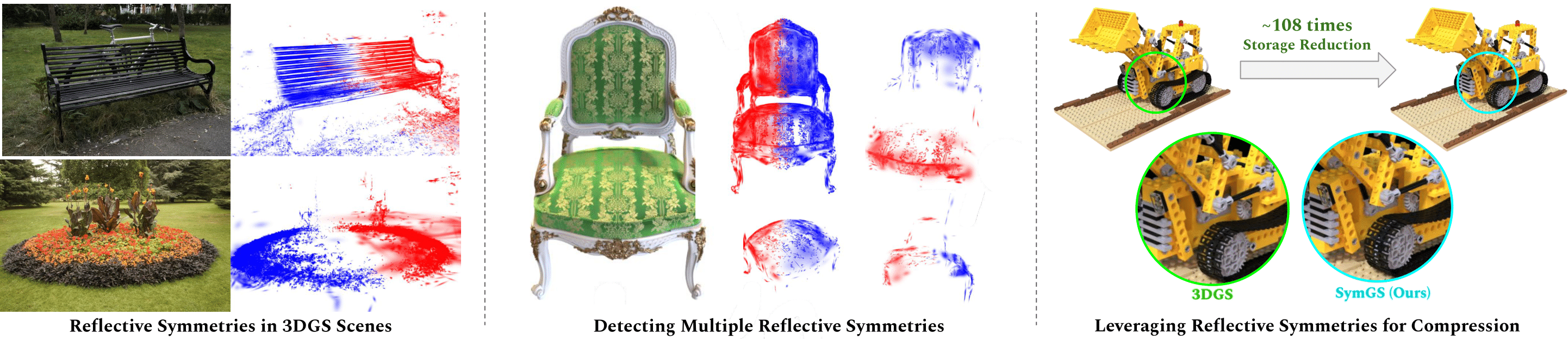}
    \captionof{figure}{Our SymGS leverages \textit{Reflective Symmetries} in a 3DGS scene for compression while preserving rendering quality.}
    \label{fig:teaser}
\end{center}
}]
\makeatletter
\newcommand{\blfootnote}[1]{%
  \begingroup
    \renewcommand\@makefntext[1]{##1}
    \renewcommand\@makefnmark{}
    \addtocounter{footnote}{1}
    \footnotetext{#1}
    \addtocounter{footnote}{-1}
  \endgroup
}
\makeatother

\blfootnote{*These authors contributed equally.}
\blfootnote{Copyright © 2026, Association for the Advancement of Artificial
Intelligence (www.aaai.org). All rights reserved.}

\begin{abstract}
  3D Gaussian Splatting has emerged as a transformative technique in novel view synthesis, primarily due to its high rendering speed and photorealistic fidelity. However, its memory footprint scales rapidly with scene complexity, often reaching several gigabytes. Existing methods address this issue by introducing compression strategies that exploit primitive-level redundancy through similarity detection and quantization. We aim to surpass the compression limits of such methods by incorporating symmetry-aware techniques, specifically targeting mirror symmetries to eliminate redundant primitives. We propose a novel compression framework, \textbf{\textit{SymGS}}, introducing learnable mirrors into the scene, thereby eliminating local and global reflective redundancies for compression. Our framework functions as a plug-and-play enhancement to state-of-the-art compression methods, (e.g. HAC) to achieve further compression. Compared to HAC, we achieve $1.66 \times$ compression across benchmark datasets (upto $3\times$ on large-scale scenes). On an average, SymGS enables $\bf{108\times}$ compression of a 3DGS scene, while preserving rendering quality. The project page and supplementary can be found at \textbf{\color{cyan}{symgs.github.io}}
\end{abstract}



\section{Introduction}


3D Gaussian Splatting (3DGS) \cite{kerbl2023gaussian} has emerged as a revolutionary technology for novel-view synthesis, offering high rendering quality and speed. However, achieving these high-quality scenes represented using 3DGS often require substantial memory, as rendering fidelity typically increases with the number of Gaussians. Large-scale scenes frequently exceed several \textit{Gigabytes}. Addressing this memory overhead is critical to enabling a broader deployment of 3DGS, particularly in resource-constrained environments such as robotics, web-based 3D applicatitons, and real-time game engines, where storage and runtime efficiency are essential. 
Despite recent advances, many approaches still generate a large number of redundant Gaussians tailored to individual views. Scaffold-GS \cite{scaffoldgs} mitigates this by introducing a sparse set of anchor points, each parameterizing a fixed local Gaussian ensemble via a learned feature vector. This formulation enables compact scene representation while preserving fidelity, further enhanced through anchor growing and pruning strategies. HAC \cite{hac2024} extends this framework by identifying and compressing redundancies in anchor features. It leverages a hash grid to learn value distributions across attributes, followed by an adaptive quantization step for discretization. Nevertheless, compression performance saturates as residual redundancies in anchor attributes remain insufficiently exploited. Additionally, there is a lack of interpretability in the aforementioned techniques, as it is difficult to visually understand the rationale behind the removal or quantization of specific Gaussian primitives. Therefore, a new axis needs to be explored for compression of 3DGS.

\noindent
Many real-world objects like desks and bottles often contain inherent local reflective symmetries regions where one part is an approximate mirror of another. We observe that exploiting these symmetries can significantly reduce redundancy, lower storage costs, and produce more structured and interpretable scene representations. However, detecting such symmetries in a 3D Gaussian Splatting (3DGS) representation is challenging. 3DGS often suffers from multi-view noise, view-dependent shading artifacts, and uneven Gaussian densities caused by sparse or biased camera trajectories. In addition, most symmetries in real-world scenes are only approximate or partial, and the space of potential mirror planes is large, making reliable detection non-trivial.

\noindent
To this end, we introduce SymGS, a novel method for compressing 3DGS by discovering and leveraging reflective symmetries. We first estimate dominant mirror planes using a symmetry detection algorithm tailored specifically for 3D Gaussians. Once a mirror plane is identified, Gaussians on one side are retained, while their counterparts are reflected across the plane and the originals discarded. The mirror plane and the retained Gaussians are then jointly optimized via differentiable splatting to minimize photometric loss to maintain visual fidelity. We apply this procedure recursively, re-running symmetry detection on the remaining Gaussians after each compression step to progressively uncover additional symmetric structures in a global-to-local fashion. As illustrated in Fig. \ref{fig:teaser}, our proposed hierarchical optimization strategy enables $108\times$ compression of the 3DGS scene while maintaining rendering quality, while also yielding a more interpretable scene decomposition. 
Additionally, we demonstrate the plug-and-play capability of our framework by integrating it into the current SOTA 3DGS compression method, HAC. We achieve superior compression rates across standard publicly available datasets, ranging from small single-object scenes to large-scale environments. 
In summary, our contributions are as follows:
\begin{itemize}
    \item We introduce a novel way of compressing 3DGS scenes by utilizing reflective symmetries in a scene to reduce the number of Gaussians.
    \item We design a CUDA-accelerated, grid-based symmetry detection algorithm tailored for 3DGS, enabling efficient detection of local symmetries on modern GPUs.
    \item We improve upon the storage efficiency of HAC by incorporating the proposed symmetry-aware framework as a plugin module, achieving SOTA compression rate.
\end{itemize}
\begin{figure}[h!]
    \centering
        \includegraphics[width=0.9\linewidth]{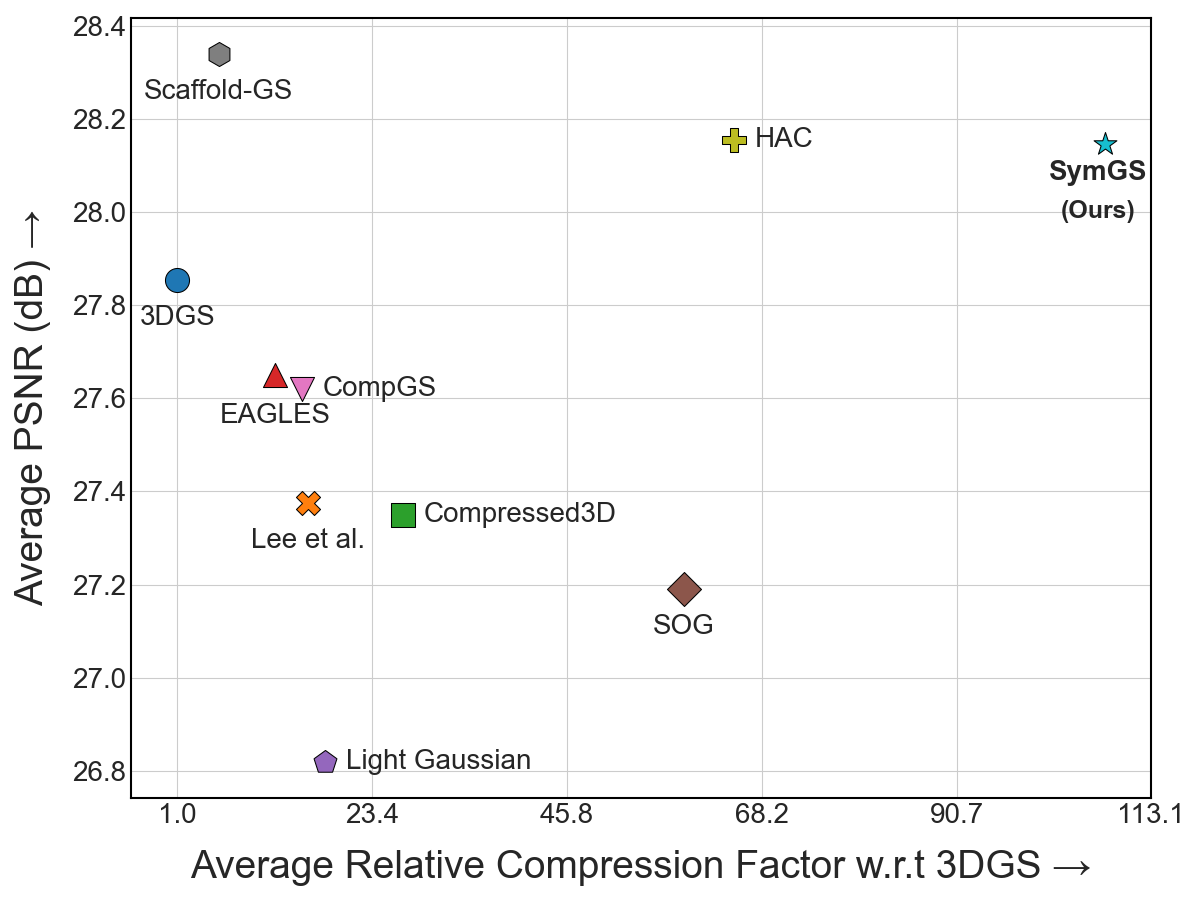}
    \caption{\textbf{PSNR vs RCF for 3DGS compression.}}
    \label{fig:comparison}
\end{figure}

\section{Related Works}
\subsection{3DGS Compression}
Many works have tried to address the problem of large storage space required by 3DGS scenes. LightGaussian \cite{lightgaussian} applies vector quantization (VQ) to selectively compress SH coefficients guided by a significance score, while RDO-Gaussian \cite{xie2024rdo} employs rate-distortion optimized quantization and entropy coding. Compact3D \cite{compact3d} proposes sensitivity-aware VQ and entropy modeling. Compact3DGS \cite{compact3dgs} replaces SH colors with hash-grid-MLP decoding, while Self-Organizing Gaussians \cite{sog} arrange primitives on 2D grids to exploit local smoothness for image-like compression. EAGLES \cite{eagles} encodes attributes as latent vectors decoded by MLPs, maintaining compactness while preserving quality. ContextGS \cite{contextgs} builds on Scaffold-GS \cite{scaffoldgs} by hierarchically predicting anchors across levels. Methods such as Octree-GS \cite{ren2024octreegsconsistentrealtimerendering} utilize depth maps and level-of-detail octrees for compaction. Among these, HAC \cite{hac2024} stands out by integrating anchor-based representation, hash-grid assisted context modeling, and adaptive quantization to achieve superior compression ratios and rendered quality across diverse datasets. Recently, HAC++ \cite{chen2025hac++} extends HAC by capturing intra-anchor contextual relationships to further enhance compression performance.

\subsection{3D Symmetry Detection}
Symmetry detection in 3D point clouds is a fundamental problem with applications in geometry processing, segmentation, reconstruction, and compression. A seminal work by Mitra et al.~\cite{mitra2006partial} introduced partial symmetry detection in point clouds via a Hough-transform-like voting scheme, where points with similar local structure contribute to a 7D accumulator representing rotation and translation parameters. While effective, this method is limited to point sets and suffers from discretization artifacts in mirror estimation. Other methods have approached this task through model-driven algorithms that detect global or partial symmetries using descriptors like generalized moments \cite{martinet2006accurate}, planar reflective transforms \cite{podolak2006planar}, and symmetry factored embeddings \cite{lipman2010symmetry}. These methods often assume clean geometry and are sensitive to noise or partial data. To overcome such limitations, learning-based approaches have emerged. PRS-Net \cite{gao2020prsnet} and E3Sym \cite{li2023e3sym} employ supervised and unsupervised deep networks, respectively, to detect global planar symmetries. The most recent progress in partial extrinsic symmetry detection is SymCL and SymML \cite{kobsik2023partial}, which leverage geodesic patches and contrastive learning to produce transformation-invariant features without requiring labels. Despite such progress, no work to date incorporates symmetry priors into 3DGS. Given the inherent symmetry in most man-made environments \cite{mitra2013symmetry}, symmetry-aware compression holds the potential to significantly improve efficiency and interpretability. This presents an open and impactful direction for extending symmetry detection methods to 3DGS.

\begin{figure*}
    \centering
    \includegraphics[width=0.9\linewidth]{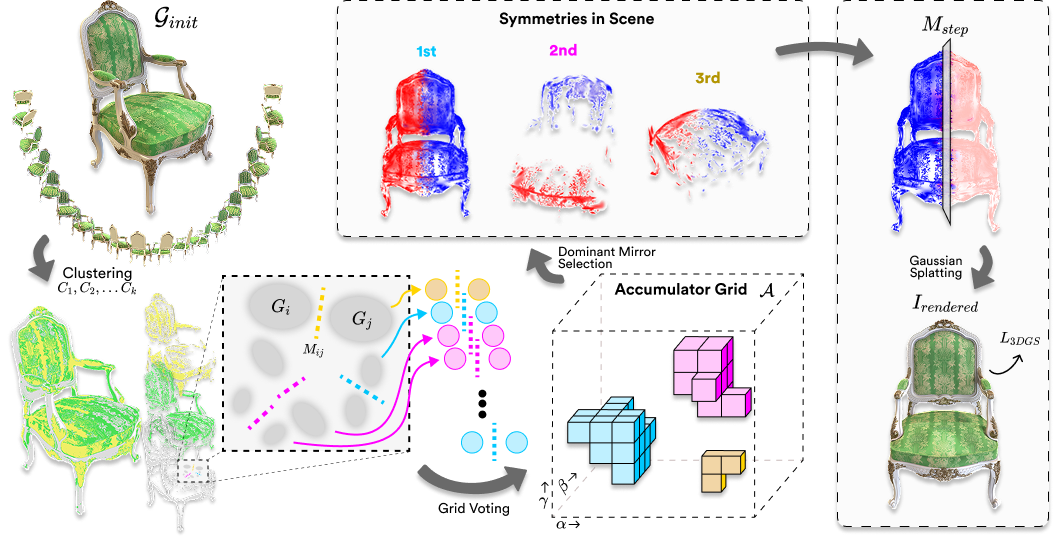}
    \caption{\textbf{SymGS Framework:}
    Given a 3DGS scene, we first perform Gaussians clustering and voting to identify the dominant mirror symmetry. Then, the Gaussians on one side of the mirror are replaced with reflections of their counterparts. Finally, the modified Gaussian set and mirror parameters are jointly optimized using the standard photometric loss.}
    \label{fig:pipeline}
\end{figure*}

\section{Methodology}
\begin{figure}[h!]
    \centering
    \includegraphics[width=1\linewidth]{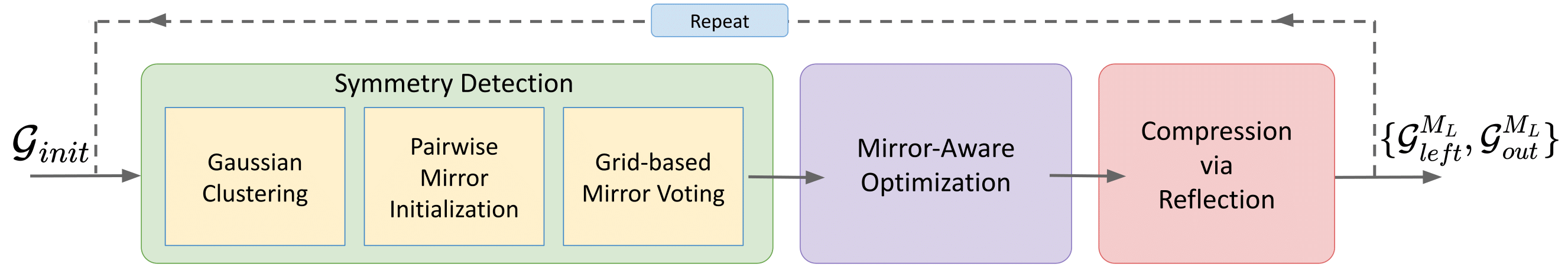}
    \caption{\textbf{Iterative Symmetry Detection \& Optimization.}}
    \label{fig:flow}
\end{figure}

Our aim is to estimate mirror planes to model reflective symmetries at different scales and leverage them for compressing the 3DGS scene. We achieve this through an iterative process, as illustrated in Fig. \ref{fig:flow}. First, we detect the most dominant symmetry in the 3DGS scene. Inspired by \cite{mitra2006partial}, we define a 3D accumulator grid - a discrete parameter space where each cell corresponds to a candidate mirror plane. Every possible pair of \textit{similar-looking} Gaussians vote for the plane that lies between them, and votes from all pairs are accumulated in the grid. The cell with the largest number of votes corresponds to the most prominent symmetry present in the scene. We start with the highest voted symmetry as initial mirror, $M_0$. Gaussians on one side of $M_0$ are discarded, and the others are reflected across it. This set of Gaussians (original and reflected) is then jointly optimized with the mirror $M_0$ via the rendering loss. Thus, the process alternates between \textbf{\textit{Symmetry Detection}} and \textbf{\textit{Mirror-Aware Optimization}}. This is applied recursively to the remaining set of Gaussians, discovering additional mirror planes $M_1, M_2, M_3, \dots$, progressively compressing the scene. Fig. \ref{fig:pipeline} shows the illustration of the SymGS framework.

\subsection{Symmetry Detection for 3DGS}
Given the initial set of Gaussians $\mathcal{G}_{init}$, we aim to estimate the most dominant mirror plane, via a clustering-based voting strategy explained as follows:
\\
\\
\noindent
\textbf{3D Gaussian Clustering:}
\label{sec:clustering}
For symmetry detection, we require clusters of \textit{similar-looking} Gaussians to avoid spurious symmetry detection. Hence, it is crucial to ensure that only visually and geometrically similar Gaussians are paired together. Therefore, all the Gaussians with attributes in the same range are clustered together. Specifically, for a Gaussian $G_i$ we define clustering based on color $c_i$, opacity $o_i$ \& scale $s_i$. We transform color $c_i$ in the HSV space and discretize each channel (H/S/V) into $\mathcal{N}_c$ bins. Opacity $o_i$ is divided into $\mathcal{N}_o$ bins, while scale $s_i$ is discretized into $\mathcal{N}_s$ bins uniformly across the x, y, and z axes. This results in a total of $\mathcal{N}_{c_H} \times \mathcal{N}_{c_S} \times \mathcal{N}_{c_V} \times \mathcal{N}_c \times \mathcal{N}_o \times \mathcal{N}_{s_x} \times \mathcal{N}_{s_y} \times \mathcal{N}_{s_z} =\mathcal{N}_{clstr}$ clusters. A Gaussian $G_i$ belonging to a cluster $C_k$, will be paired with every other Gaussian $G_j\in C_k$.
\begin{figure*}[h!]
\centering
    \includegraphics[width=\linewidth]{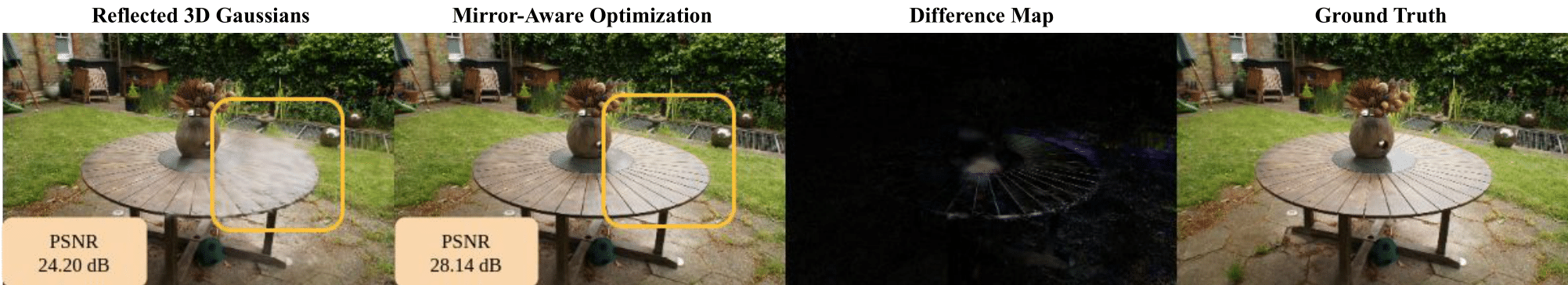}
\caption{Mirror-aware optimization restores the visual details which might get attenuated while reflecting the Gaussians.}
\label{fig:mirror_opt}
\end{figure*}
\\
\\
\noindent
\textbf{Mirror Parametrization:}
\label{sec:detection}
Any \textit{mirror} or plane of reflective symmetry can be characterized by the general equation of a 3D plane, i.e. $a x + b y + c z = 1.$
However, the parameters $a$, $b$, and $c$ are individually unbounded for a given scene, making exhaustive search in this space impractical. To define a finite and tractable set of candidate mirrors, we reparameterize the plane using a bounded set of parameters ($\alpha,\beta,\gamma$). Specifically, we express the plane normal in spherical coordinates with angular parameters $\alpha \in [0, \pi]$ and $\beta \in [0, 2\pi]$, and define the signed distance from the origin as $\gamma \in [$0$, \text{\textit{extent}}]$, where \textit{extent} denotes the radius of the scene.
%
While this polar representation bounds the mirror plane parameter space, it remains continuous and infinitely dense. To make symmetry detection computationally feasible, we discretize this space into a uniform 3D accumulator voxel-grid $\mathcal{A}$, defined over $(\alpha, \beta, \gamma)$ with dimension $d_\alpha \times d_\beta \times d_\gamma$, where:
%
%
\[
d_\alpha = \lfloor \frac{\pi}{\alpha_{res}} \rfloor, \quad  
d_\beta = \lfloor \frac{2\pi}{\beta_{res}} \rfloor, \quad  
d_\gamma = \lfloor \frac{\text{\textit{extent}}}{\gamma_{res}} \rfloor
\]  
\\
and, $\alpha_{res}, \beta_{res}, \gamma_{res}$ are voxel size along the three axes respectively.
Each voxel $v\in \mathcal{A}$ represents a possible mirror in the 3D scene, and its integer value denotes how many Gaussian pairs are associated with that mirror.\\
\noindent
 For a pair of Gaussians $(G_i, G_j) \in C_k$, and their respective mean 3D positions $\text{\textbf{x}}_i$ and $\text{\textbf{x}}_j$, we initialize a mirror plane $\mathcal{M}_{ij}$ passing between them, with center $\textbf{c}$ and normal $\textbf{n}$ given as:
\[
\text{\textbf{c}} = \frac{(\text{\textbf{x}}_i + \text{\textbf{x}}_j)}{2} \quad ,  \quad 
\text{\textbf{n}} = \frac{(\text{\textbf{x}}_i - \text{\textbf{x}}_j)}{|(\text{\textbf{x}}_i - \text{\textbf{x}}_j)|}
\]
We compute the parameters ($\alpha,\beta,\gamma$) for $M_{ij}$ as:
\[
\alpha =cos^{-1}\textbf{n}_x, \quad
\beta = \pi + tan^{-1}\frac{\textbf{n}_y}{\textbf{n}_x}, \quad
\gamma = \textbf{n} \cdot \textbf{c}
\]
The 3D index of the voxel $v_{ij}\in \mathcal{A}$ representing mirror $\mathcal{M}_{ij}$ is then computed as:
\[
v^{\alpha}_{ij} = \lfloor \frac{\alpha}{\alpha_{res}} \rfloor,  \quad
v^{\beta}_{ij} =\lfloor \frac{\beta}{\beta_{res}} \rfloor,  \quad
v^{\gamma}_{ij} =\lfloor \frac{\gamma}{\gamma_{res}} \rfloor.
\]
All the voxels of the accumulator grid are initialized as zero, i.e. $\mathcal{A}[v^\alpha_{ij}][v^\beta_{ij}][v^\gamma_{ij}]=0$ for each $v_{ij}\in \mathcal{A}$.
\\
\\
\noindent
\textbf{Grid-based Mirror Voting:}
Initially, all possible pairs of similar Gaussians $(G_i,G_j) \in C_k$ define a set of \textit{candidate mirrors}, out of which, the one exhibiting the most dominant reflective symmetry is selected. A vote is cast by a pair of similar Gaussians for the mirror plane that best explains their reflective symmetry. In other words, for the pair of Gaussians $(G_i,G_j)$, the value of the voxel $v_{ij}$, i.e. $\mathcal{A}[v^\alpha_{ij}][v^\beta_{ij}][v^\gamma_{ij}]$ is incremented by one. After accumulating votes from all possible Gaussian pairs for all clusters into the common $\mathcal{A}$, the voxel with the highest vote count represents the most dominant mirror plane, ${M}_0$.
The dominant mirror plane $M_0$, divides the symmetric subpart of the scene into two halves, $G_{left}^{M_0}$ (in the direction of normal) and $G_{right}^{M_0}$ (opposite to the normal direction), consisting of all the Gaussian pairs that contributed to its votes. The remaining pairs of Gaussians which do not vote for the mirror $M_0$ are outside of its influence, denoted as $G^{M_0}_{out}$.\\
\\
\noindent
The symmetry detection algorithm, with \(O(n^2)\) complexity, is critical for runtime efficiency. To simultaneously handle millions of Gaussians in large scenes, we design a CUDA-accelerated variant, reducing the time to identify the largest mirror from tens of minutes to around $10$ seconds for roughly $1$ million Gaussians.
All the pairs of 3D Gaussians are assigned individual GPU threads, parallely casting their votes to the shared accumulator grid in an atomic operation. 
\textit{Please refer to the supplementary for timing analysis for different implementation choices.}
\begin{figure*}[h!]
    \centering
    \includegraphics[trim={0 0.2cm 0 0},clip, width=1\linewidth]{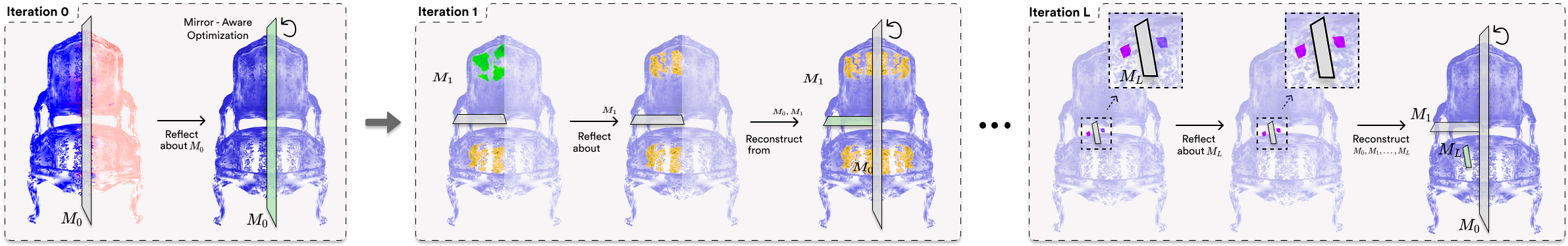}
    \caption{At each iteration step, the Gaussians in $G^{M_{step}}_{left}$ replace their symmetrical counterparts in $G^{M_{step}}_{right}$. The full scene is reconstructed by iteratively reflecting these Gaussians across all previous mirror levels in succession.}
    \label{fig:optimization}
\end{figure*}
\subsection{Mirror-Aware Optimization}
\label{sec:optimization}
After the initial estimation of $M_0$, we discard $G^{M_0}_{right}$ and replace it by reflecting the Gaussians $G^{M_0}_{left}$. However, this leads to a significant degradation in rendering quality (see Fig. \ref{fig:mirror_opt}). This is primarily due to the finite number of thresholds involved in the clustering process, and the discrete approximation of true mirror parameters owing to the finite resolution of the accumulator grid. 

To account for such inconsistencies, we propose to make the mirror plane optimizable by allowing its normal and center to adjust during training. Given the parameters of $M_0$ and the attributes of Gaussians $G^{M_0}_{left}$, during forward pass we construct the scene by first reflecting the $G^{M_0}_{left}$ across $M_0$ to generate the symmetrical part of the scene, and then adding the rest of the Gaussians $G^{M_0}_{out}$ into the scene. The reflection involves mirroring the position of $G^{M_0}_{left}$ about $M_0$, while copying the scale and opacity as it is. The final set of Gaussians are then passed to the differential 3DGS rasterizer to obtain the rendered image as per viewing direction. Finally, the rendered image is compared with the ground truth (input) using the standard 3DGS rendering loss \cite{kerbl20233d}. During the backwards pass, the gradient updates are applied to attributes of $G^{M_0}_{left}$ \& $G^{M_0}_{out}$, and to the parameters of $M_0$ as well. This joint-optimization of the mirror plane and Gaussian attributes makes it possible to refine the symmetry that best aligns with the corresponding sub-structure of the scene, while correcting for discretization errors and inconsistencies involved in initial mirror estimates.

\subsection{Compression via Reflection}
\label{sec:compression}
By symmetric duplication of Gaussians $G^{M_0}_{left}$, discarding Gaussians $G^{M_0}_{right}$, we effectively halve the number of primitives that are required to represent the symmetric part of the scene corresponding to $M_0$. Meanwhile, the remaining Gaussians $G^{M_0}_{out}$, i.e. those excluded from $M_0$, capture within it, other non-dominant symmetries, along with asymmetric or non-redundant details, such as occlusions, clutter, or scene-specific irregularities. Only the mean positions of one half of the symmetric Gaussians, i.e. $X^{M_0}_{left}$ and mirror parameters ($\alpha,\beta,\gamma$) need to be stored, while the remaining subpart is reconstructed on-the-fly at training time. Please refer to Fig. \ref{fig:optimization} for a visual depiction of the reconstruction process. This reduces the overall storage requirements.
\subsection{Iterative Compression}
%
%
After optimizing the scene for the first mirror $M_0$, we focus on identifying symmetries in the remaining scene, i.e. $G^{M_0}_{left}$ and $G^{M_0}_{out}$. More generally, for any subsequent iteration `$step$', the subpart of the scene to be considered for further compression consists of Gaussian set $G^{M_{step}}$ =  \{$G^{M_{step-1}}_{left}$, $G^{M_{step-1}}_{out}$\} from previous iteration.
As illustrated in Fig. \ref{fig:flow}, we follow the same process from Gaussian clustering to mirror-aware optimization, considering only $G^{M_{step}}$, resulting into the current most dominant, optimized symmetry plane $M_{step}$. Thus, this iterative process from initial step to last step yields a set of mirrors \{$M_0, M_1, M_2, \dots, M_{L}$\}, progressively discovering similarities till level $L$. 
At every iteration, we \textbf{only} store mean positions $X^{M_{step}}_{left}$ \& parameters of $M_{step}$, discarding $G^{M_{step}}_{right}$ altogether.
%
For mirror-aware optimization at step $L$, we reconstruct the scene from the current level. We load attributes of $G^{M_{L}}_{left}$ and $G^{M_{L}}_{out}$, reflect the $G^{M_{L}}_{left}$ about $M_{L}$ to obtain $\widehat{G}^{M_{L}}_{right}$ (as it is not stored). For step $L-1$, we obtain $G^{M_{L-1}}_{left+out} = \{G^{M_{L}}_{left}, \widehat{G}^{M_{L}}_{right},G^{M_{L}}_{out}$\}. We load already stored mean positions $X^{M_{L-1}}_{left}$ and perform nearest neighbor query search within $G^{M_{L-1}}_{left+out}$, to obtain $G^{M_{L-1}}_{left}$. We then estimate $G^{M_{L-1}}_{out} = G^{M_{L-1}}_{left+out} - G^{M_{L-1}}_{left}$, and further reflect $G^{M_{L-1}}_{left}$ to obtain $\widehat{G}^{M_{L-1}}_{right
}$. We follow the same process recursively for further lower levels, continuing this reconstruction to obtain the complete scene, which is represented by $G^{M_{0}}_{left+out} = \{G^{M_{0}}_{left}, \widehat{G}^{M_{0}}_{right},G^{M_{0}}_{out}$\}. We then apply mirror-aware optimization, tuning only the mirror $M_{L}$. The aforementioned hierarchical compression strategy, where each mirror $M_{step}$ incrementally reduces the number of Gaussians, achieves coarse-to-fine structure-aware compression across the full scene. Finally, we store only $X^{M_{l}}_{left}$ and mirror parameters $M_l$ for each level $l$ (denoted collectively as $X_{ret}$ and $M$ respectively), as well as the last level attributes $G^{M_{L}}_{left}$ and $G^{M_{L}}_{out}$. \textit{Please find pseudocode in supplementary}

\definecolor{lightred}{rgb}{1, 0.8, 0.8}
\definecolor{lightyellow}{rgb}{1, 1, 0.7}

\begin{table*}[t!]
  \centering
  \caption{SymGS achieves higher Relative Compression Rate (RCF) w.r.t. 3DGS, while maintaining the comparable PSNR \colorbox{lightred}{Red} and \colorbox{lightyellow}{yellow} highlight respectively denote the best and second best metrics among the compression methods.}
  \label{tab:quantitative}
  \resizebox{\textwidth}{!}{%
    \centering
    \centering
    \begin{tabular}{l|ccc|ccc|ccc|ccc|ccc}
      \toprule
      \textbf{Datasets} & \multicolumn{3}{c|}{\textbf{Synthetic-NeRF}} & \multicolumn{3}{c|}{\textbf{Mip-NeRF360}} & \multicolumn{3}{c|}{\textbf{Tank \& Temples}} & \multicolumn{3}{c|}{\textbf{Deep Blending}} & \multicolumn{3}{c}{\textbf{BungeeNerf}} \\
      \textbf{Methods}& PSNR $\uparrow$ & Size $\downarrow$ & RCF $\uparrow$ & PSNR $\uparrow$ & Size $\downarrow$ & RCF $\uparrow$ & PSNR $\uparrow$ & Size $\downarrow$ & RCF $\uparrow$ & PSNR $\uparrow$ & Size $\downarrow$ & RCF $\uparrow$& PSNR $\uparrow$ & Size $\downarrow$ & RCF $\uparrow$\\
      \midrule
      3DGS & 33.80 & 68.46 & 1.00$\times$ & 27.49 & 744.7 & 1.00$\times$ & 23.69 & 431.0 & 1.00$\times$ & 29.42 & 663.9 & 1.00$\times$ & 24.87 & 1616 & 1.00$\times$\\
      \midrule
      Lee et al. & \cellcolor{lightyellow}33.33 & 5.54 & 12.35$\times$ & 27.08 & 48.80 & 15.26$\times$ & 23.32 & 39.43 & 10.93$\times$ & 29.79 & 43.21 & 15.36$\times$ & 23.36 & 82.60 & 19.56$\times$ \\
      Compressed3D & 32.94 & 3.68 & 18.60$\times$ & 26.98 & 28.80 & 25.85$\times$ & 23.32 & 17.28 & 24.94$\times$ & 29.38 & 25.30 & 26.24$\times$ & 24.13 & 55.79 & 28.96$\times$ \\
      EAGLES & 32.54 & 5.74 & 11.92$\times$ & 27.15 & 68.89 & 10.80$\times$ & 23.41 & 34.00 & 12.67$\times$ & 29.91 & 62.00 & 10.71$\times$ & 25.24 & 117.1& 13.80$\times$ \\
      Light Gaussian & 32.73 & 7.84 & 8.73$\times$ & 27.00 & 44.54 & 16.71$\times$ & 22.83 & 22.43 & 19.21$\times$ & 27.01 & 33.94 & 19.56$\times$ & 24.52 & 87.28 & 18.51$\times$ \\
      SOG & 31.05 & 2.20 & 31.11$\times$ & 26.01 & 23.90 & 31.15$\times$ & 22.78 & 13.05 & 33.02$\times$ & 28.92 & 8.40 &79.03$\times$ &  - & - & - \\
      CompGS & 33.09 & 4.42 & 15.48$\times$ & 27.16 & 50.30 & 14.80$\times$ & 23.47 & 27.97 & 15.40$\times$ & 29.75 & 42.77 & 15.52$\times$ & 24.63 & 104.3 & 15.49$\times$ \\
      Scaffold-GS & \cellcolor{lightred}33.41 & 19.36 & 3.53$\times$ & \cellcolor{lightred}27.50 & 253.9 & 2.93$\times$ & 23.96 & 86.50 & 4.98$\times$ & \cellcolor{lightred}30.21 & 66.00 & 10.05$\times$ & \cellcolor{lightred}26.62 & 183.0 & 8.83$\times$ \\
      HAC & 33.05 & \cellcolor{lightyellow}1.38 & \cellcolor{lightyellow}49.60$\times$ & \cellcolor{lightyellow}27.29 & \cellcolor{lightyellow}16.66 & \cellcolor{lightyellow}44.69$\times$ & \cellcolor{lightred}24.14 & \cellcolor{lightyellow}8.07 & \cellcolor{lightyellow}53.40$\times$ & 29.85 & \cellcolor{lightyellow}7.80 & \cellcolor{lightyellow}85.11$\times$ & \cellcolor{lightyellow}26.44 & \cellcolor{lightyellow}20.28 & \cellcolor{lightyellow}79.68$\times$ \\
      \textbf{SymGS (Ours)} & 33.19 & \cellcolor{lightred}0.95 & \cellcolor{lightred}{72.06$\times$} & \cellcolor{lightyellow}27.29 & \cellcolor{lightred}11.74 & \cellcolor{lightred}{63.43$\times$} & \cellcolor{lightyellow}24.02 & \cellcolor{lightred}7.14 & \cellcolor{lightred}{60.36$\times$} & \cellcolor{lightyellow}29.99 & \cellcolor{lightred}2.59 & \cellcolor{lightred}{256.33$\times$} & 26.24 & \cellcolor{lightred}10.29 & \cellcolor{lightred}{157.05$\times$} \\
      \bottomrule
    \end{tabular}
    }


\end{table*}

\subsection{Integration with HAC}
Our framework for finding and exploiting reflective symmetries to reduce the number of primitives can be applied to other existing compression methods. To showcase this, we build on top of HAC \cite{hac2024}, a recent SOTA in 3DGS compression, integrating symmetry detection and progressive compression within its pipeline. HAC achieves good compression rate mainly through its adaptive quantization module and a hash-grid to estimate the quantization steps for the \textbf{\textit{anchor}} attributes. We apply the propsed iterative symmetry detection and optimization process over these \textit{anchors}. For a given 3DGS scene, we first run HAC to estimate anchor attributes, and then compute the average color, opacity, and scale for each anchor by aggregating over its $k$ associated Gaussians across all views. These aggregated attributes are then used to cluster the anchors, which is analogous to the Gaussian clustering process. We then perform grid-based voting to identify the most dominant mirror plane at all levels. For each level, we then optimize the mirror and anchor features, hash-grid, and all the MLPs involved in HAC. 
%
%
We then iteratively compress the scene by following the same process described in the previous section.
\textit{Please refer to the supplementary for more details.}

\begin{figure*}[h!]
    \centering
    \includegraphics[width=\linewidth]{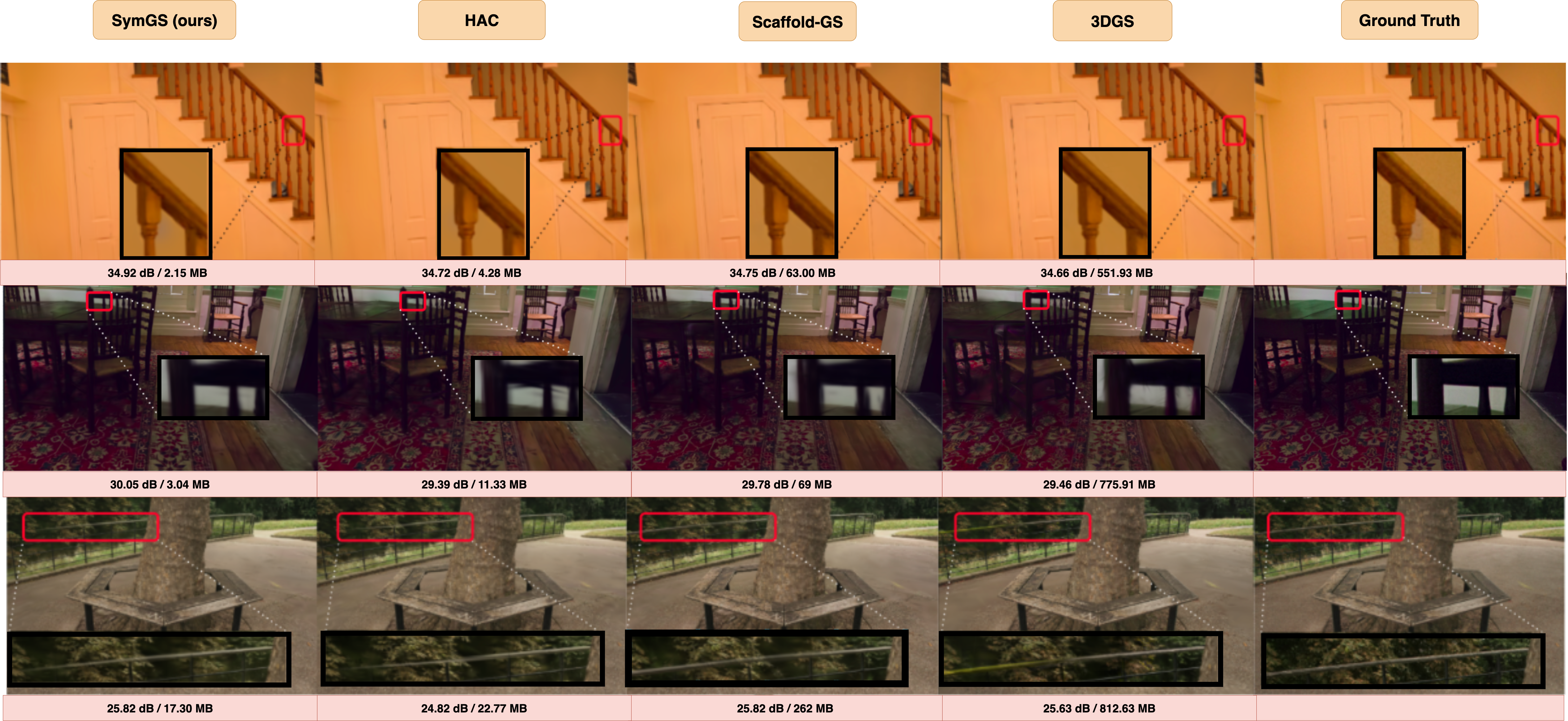}
    \caption{\textbf{Qualitative Comparison with HAC, Scaffold-GS and 3DGS}.}
    \label{fig:qualitative}
\end{figure*}

\section{Experiments \& Results}
\subsubsection{Datasets:}

%
We evaluate our method on the five standard benchmarking datasets as in HAC, conducting experiments across a total of 27 scenes: 8 from Synthetic-NeRF \cite{mildenhall2021nerf}, all 9 scenes from Mip-NeRF360 \cite{Barron_2022_CVPR}, 2 from Tanks \& Temples \cite{Knapitsch2017}, 2 from DeepBlending \cite{DeepBlending2018}, and 6 from BungeeNeRF \cite{xiangli2023bungeenerfprogressiveneuralradiance}. This selection allows us to demonstrate compression performance on both large-scale scenes, such as those in BungeeNeRF, and smaller, object-centric scenes, such as those in Synthetic-NeRF. Collectively, these datasets span a diverse range of indoor and outdoor environments.
\subsubsection{Implementation \& Training Details:}

We build upon the HAC \cite{hac2024} framework written in PyTorch, trained on an NVIDIA RTX A6000 GPU. We keep the mirror parameters $\alpha$, $\beta$ and $d$ trainable with learning rates $1e-3$ each. We choose the distance spacing $\gamma_{res}$ according to the extent of the scene. For Synthetic-NeRF and BungeeNeRF, $\gamma_{res} = 0.01$ and for MIP, TandT and DeepBlending, we take $\gamma_{res} = 0.1$. $\alpha_{res}$ and $\beta_{res}$ are fixed to $0.01$.
%
%
%

\subsection{Comparisons}
%
%
We compare with 3DGS \cite{kerbl20233d} and other existing compression methods \cite{lee2024compact}, Compressed3D \cite{niedermayr2024compressed}, EAGLES \cite{girish2024eagles}, Light Gaussian \cite{fan2025lightgaussian}, Self-Organizing Gaussian Grids (SOG) \cite{morgenstern2024compact}, CompGS \cite{navaneet2024compgs}, Scaffold-GS \cite{scaffoldgs} and HAC \cite{hac2024}. We use HAC as our primary baseline, and extend it with our symmetry detection and mirror-aware optimization.\\
\noindent
We report the final compressed model size (in MB) along with PSNR, evaluated after decoding, comparing with existing SOTA methods in Table \ref{tab:quantitative}. Across the five benchmark datasets, our approach achieves an average Relative Compression Factor (RCF) of $108 \times$ compared to 3DGS. In comparison to current SOTA HAC, we achieve $1.66 \times$ compression on an average. Specifically, we observe $3.01 \times$ compression on Deep Blending, $1.97 \times$ on BungeeNeRF, $1.4 \times$ on Mip-NeRF360, $1.4 \times$ on Synthetic-NeRF, and $1.1 \times$ on Tanks and Temples, while maintaining visual fidelity.\\
\noindent
We show qualitative comparison of SymGS with other methods in Fig. \ref{fig:qualitative}. 
We highlight the detail-preserving compression capability of SymGS, which are often lost by quantization-based approaches. 
\noindent
\subsection{Ablations}


\subsubsection{Effect of Accumulator Grid Resolution:}
\begin{figure}[h]
    \includegraphics[width=1\columnwidth]{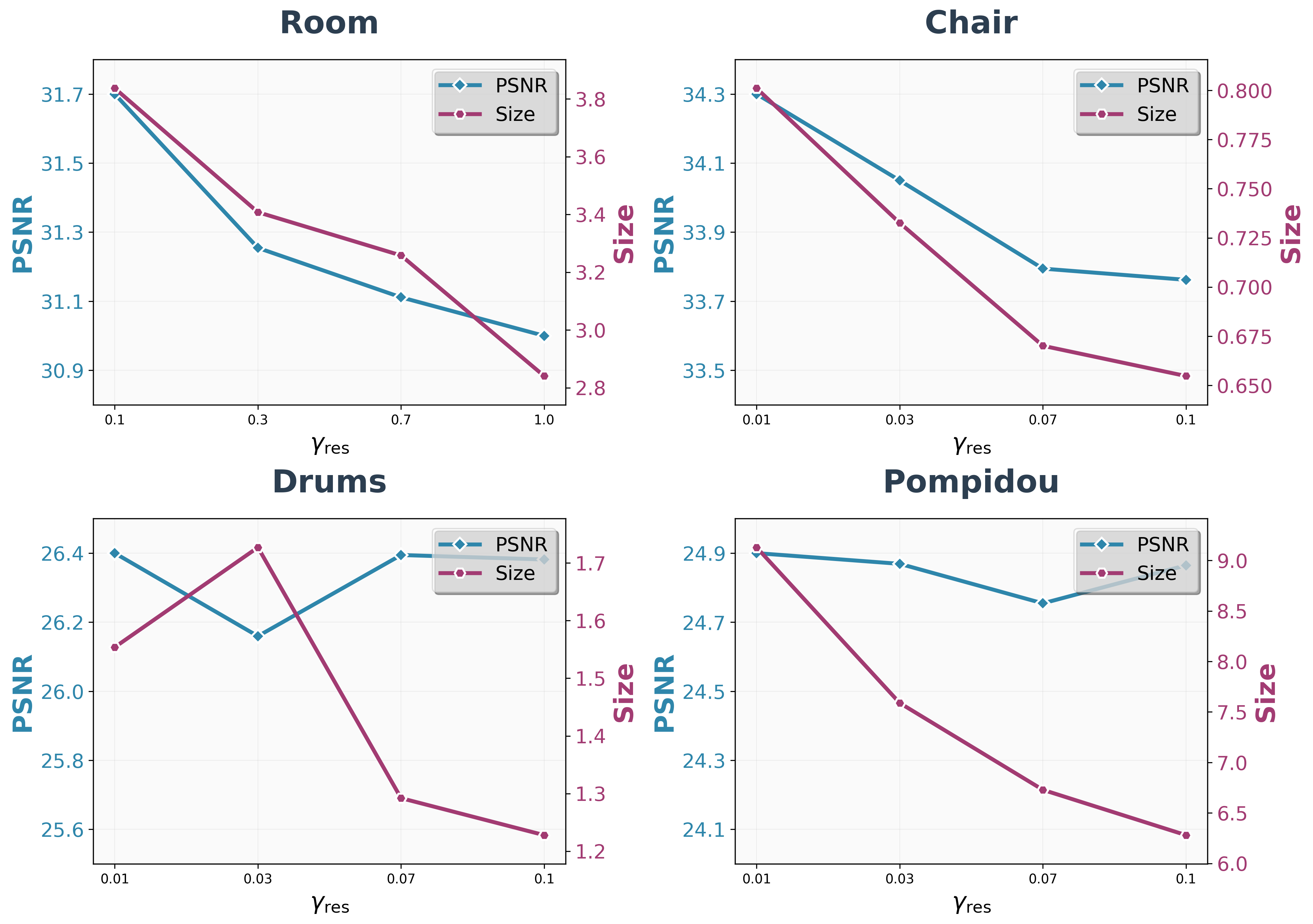}
    \caption{Effect of $\gamma_{res}$ on the PSNR (db) storage size (MB).}
    \label{fig:abl_dis}
\end{figure}

As previously mentioned, the dimensionality of the Voxel Grid $\mathcal{A}$, is dependent on the radius or extent of the scene, parametrized by $\gamma_{res}$, which  controls the level of discretization along this axis. We analyze the effect of $\gamma_{res}$ by varying it along a range based on the extent of the scene - 0.01 and 0.1 for Nerf-Synthetic scenes due to their small extent, and 0.1 to 1 for MIP scenes having a larger extent. A lower value of $\gamma_{res}$ corresponds to a more fine-grained discretization, and vice versa. A coarse discretization (large value of $\gamma_{res}$) leads to a higher number of incorrect votes between anchors during symmetry detection, resulting in a lower PSNR. This also leads to incorrect pair of Gaussians voting for mirror candidates, leading to poor symmetry detection, and thereby  decreasing the compression rate. This trend can be seen in Fig. \ref{fig:abl_dis}. Please note that we don't ablate the effect of $\alpha_{res}$ \& $\beta_{res}$ as $d_{\alpha}$ they $d_{\beta}$ do not model the extent of the scene, but only the orientation of the mirrors.

\subsubsection{Effect of Hash-Grid-MLP:}
In HAC, the usage of Hash-grid MLP for adaptive quantization for the anchor attributes enable anchors to be better aligned to each scene, leading to a better compression rate. Therefore, we ablate the effect of fine-tuning the Hash-grid MLP during training. As shown in Table \ref{tab:mlp_ablation}, the rendering quality increases when finetuned.

\begin{table}[h!]
\small
\centering
\caption{Ablation Study: \textbf{Effect of Hash Grid MLP} and \textbf{Training Attributes across multiple mirrors}. Values are shown in PSNR/Size (in MB) format.}
\label{tab:mlp_ablation}
\begin{tabularx}{\columnwidth}{l |>{\centering\arraybackslash}X |>{\centering\arraybackslash}X| >{\centering\arraybackslash}X}
\toprule
\textbf{Scene} & \textbf{w/o MLP} & \textbf{(L=5) + MLP} & \textbf{(L=1) + MLP} \\
\midrule
Chair & 33.88 / 0.93 & \textbf{34.60} / 0.82 & 34.34 / \textbf{0.80} \\
Drums & 26.13 / 1.73 & \textbf{26.40} / 1.62 & 26.37 / \textbf{1.55} \\
Room  & 31.26 / 3.92 & \textbf{31.92} / 4.15 & 31.70 / \textbf{3.84} \\
\bottomrule
\end{tabularx}
\end{table}

\subsubsection{Optimizing multiple mirrors at once:}
Primarily, the only anchor parameters affected by introducing a new mirror belong to the last level, i.e. ${G}^{M_L}_{left}$ and ${G}^{M_L}_{out}$. Thus, we only train these parameters. However, letting the previous level mirror parameters to update allows lower levels to adjust to the newly added mirror as well. Thus, instead of training only the anchors participating in the most recent mirror, we train the anchors that took part in the last 5 mirror levels. This effect can be seen in Table \ref{tab:mlp_ablation}, showcasing higher PSNR values, but empirically lower compression rates.
\section{Discussion}
\begin{figure}[h!]
    \centering
    \includegraphics[width=1\linewidth]{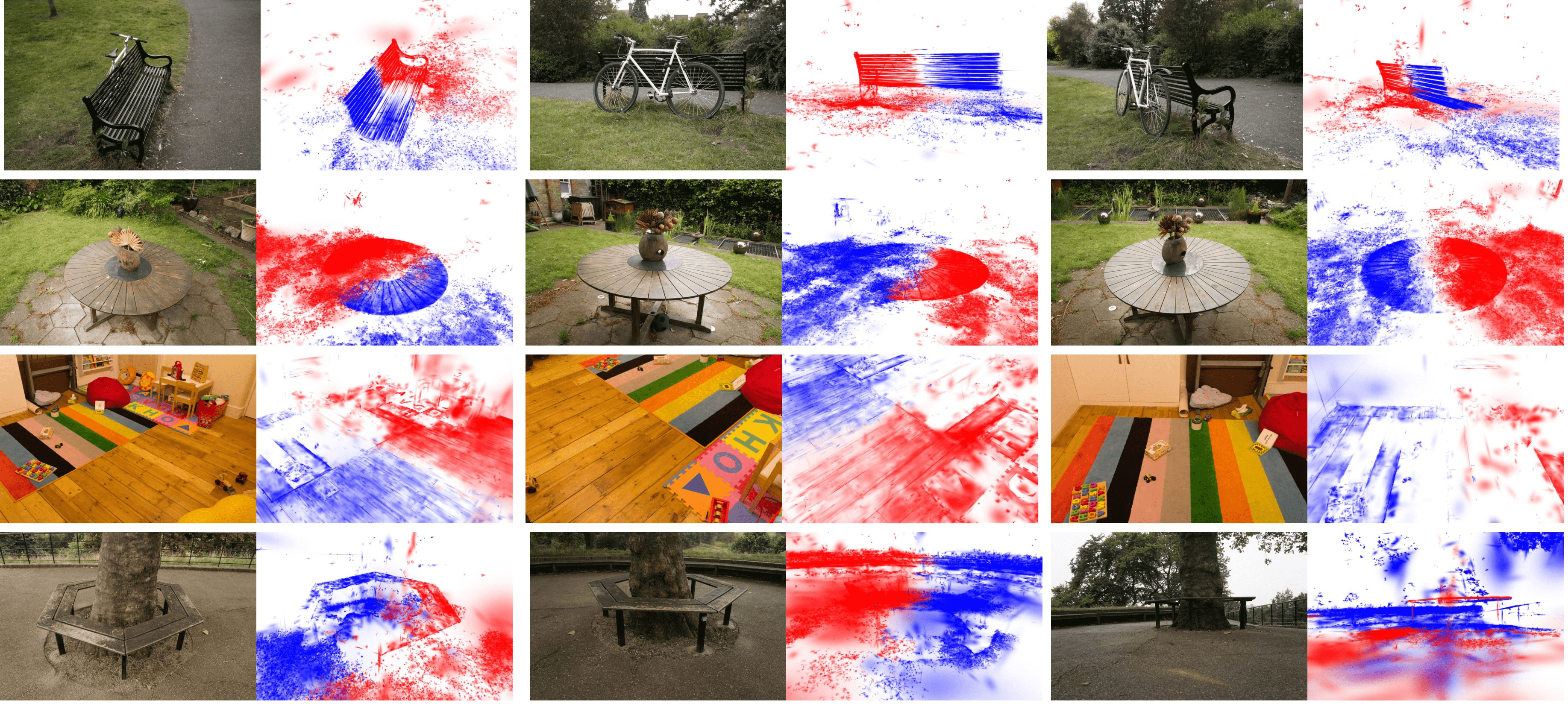}
    \caption{Detected reflective symmetries in 3DGS scenes.}
    \label{fig:symmetries3DGS}
\end{figure}

SymGS achieves superior compression gains by carefully exploiting several reflective symmetries present in a 3DGS scene (as shown in Fig. \ref{fig:symmetries3DGS}). High compression rates on BungeeNeRF dataset can be attributed to the prevalence of large-scale outdoor scenes containing numerous man-made, symmetrical structures. Similarly, Deep Blending contains indoor scenes with recurring symmetric patterns across walls, floors, and household objects, enabling further compression. The relatively low compression gain on Synthetic-NeRF is due to the fact that these scenes are already small in scale and exhibit limited geometric and appearance complexity, making them inherently more compact and less redundant.
We also observe an increase in PSNR values for Synthetic-NeRF and Deep Blending. This suggests that enforcing structural symmetry not only reduces redundancy but also enhances rendering fidelity, especially in inherently symmetric scenes such as Lego (in Synthetic-NeRF) and DrJohnson (in Deep Blending). Notably, the rendering FPS remains the same as HAC, as all the mirror decompositions are pre-computed, and no additional operations are required during rendering, making our framework highly practical. 
\section{Conclusion}
We present a novel framework for compressing 3DGS scenes by exploiting inherent reflectional symmetries found in real-world environments. By introducing a CUDA-accelerated, grid-based symmetry detection algorithm for 3DGS, we enable efficient identification and utilization of local mirror symmetries to significantly reduce redundancy in Gaussian representations. Our hierarchical optimization approach recursively uncovers and compresses symmetric structures, achieving up to 108× compression while preserving high rendering fidelity.
Furthermore, we demonstrated the versatility of SymGS by integrating it as a plug-and-play module into the state-of-the-art HAC compression method, resulting in superior compression rates across diverse datasets. This symmetry-aware compression not only improves storage efficiency but also yields more interpretable and structured scene decompositions.

\bibliography{aaai2026}

\end{document}